\documentclass[letterpaper]{article} 
\usepackage{aaai25}  
\usepackage{times}  
\usepackage{helvet}  
\usepackage{courier}  
\usepackage[hyphens]{url}  
\usepackage{graphicx} 
\usepackage{natbib}  
\usepackage{caption} 
\usepackage{algorithm}

\usepackage{enumitem}
\usepackage{diagbox}
\usepackage{amssymb}
\usepackage{amsmath}

\usepackage{tabularx}
\usepackage{multirow} 
\usepackage{booktabs}
\usepackage{arydshln}

\usepackage{newfloat}
\usepackage{listings}
\DeclareCaptionStyle{ruled}{labelfont=normalfont,labelsep=colon,strut=off} 
\floatstyle{ruled}
\newfloat{listing}{tb}{lst}{}
\floatname{listing}{Listing}
\title{ScaleOT: Privacy-utility-scalable Offsite-tuning with Dynamic LayerReplace and Selective Rank Compression}
\author{
    Kai Yao\textsuperscript{\rm 1,2},
    Zhaorui Tan\textsuperscript{\rm 3},
    Tiandi Ye\textsuperscript{\rm 4},
    Lichun Li\textsuperscript{\rm 2},\\
    Yuan Zhao\textsuperscript{\rm 2},
    Wenyan Liu\textsuperscript{\rm 2},
    Wei Wang\textsuperscript{\rm 2}\thanks{Corresponding authors.},
    Jianke Zhu\textsuperscript{\rm 1}$^*$,
}
\affiliations{
    \textsuperscript{\rm 1}Zhejiang University
    \textsuperscript{\rm 2}Ant Group
    \textsuperscript{\rm 3}University of Liverpool
    \textsuperscript{\rm 4}East China Normal University\\
    jiumo.yk@antgroup.com, jkzhu@zju.edu.cn
}

\begin{document}
\maketitle
\begin{abstract}
Offsite-tuning is a privacy-preserving method for tuning large language models (LLMs) by sharing a lossy compressed emulator from the LLM owners with data owners for downstream task tuning. This approach protects the privacy of both the model and data owners. However, current offsite tuning methods often suffer from adaptation degradation, high computational costs, and limited protection strength due to uniformly dropping LLM layers or relying on expensive knowledge distillation. To address these issues, we propose ScaleOT, a novel privacy-utility-scalable offsite-tuning framework that effectively balances privacy and utility. ScaleOT introduces a novel layerwise lossy compression algorithm that uses reinforcement learning to obtain the importance of each layer. It employs lightweight networks, termed harmonizers, to replace the raw LLM layers. By combining important original LLM layers and harmonizers in different ratios, ScaleOT generates emulators tailored for optimal performance with various model scales for enhanced privacy protection. Additionally, we present a rank reduction method to further compress the original LLM layers, significantly enhancing privacy with negligible impact on utility. Comprehensive experiments show that ScaleOT can achieve nearly lossless offsite tuning performance compared with full fine-tuning while obtaining better model privacy.
\end{abstract}

\section{Introduction}

\begin{figure}[t]
\centering
\includegraphics[width=.95\linewidth]{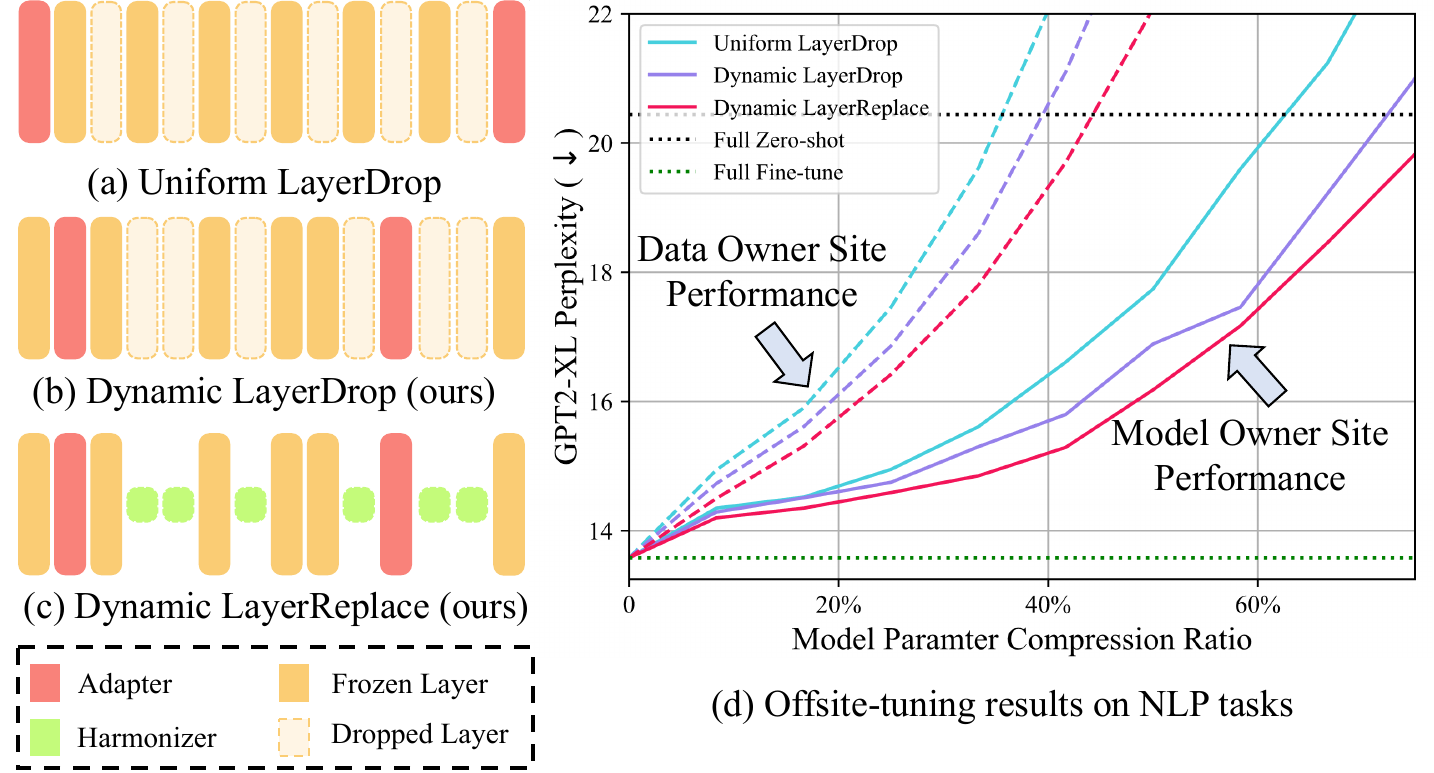}
\caption{
Comparison of layerwise compression strategies. (a)~Uniform LayerDrop. (b)~Our Dynamic LayerDrop drops layers with the estimated importance scores. (c)~Our Dynamic LayerReplace with harmonizers. (d)~Results of using different compression ratios. Our approach achieves better performance at the owner site while maintaining the performance discrepancy.
}
\label{fig:banner}
\end{figure}

\begin{figure*}[t]
\centering
\includegraphics[width=0.85\linewidth]{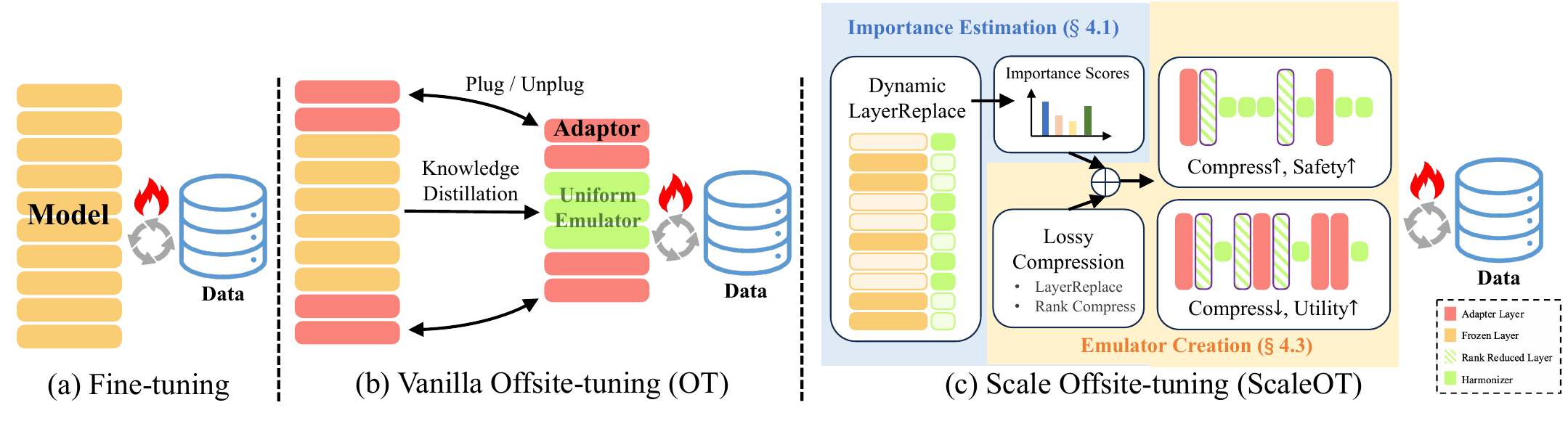}
\caption{Comparison in various tuning methods. (a)~Fine-tuning requires access to full model parameters and necessitates the co-location of data and model. (b)~Vanilla OT~\cite{OT} allows downstream users to fine-tune adapters on a lossy compressed emulator and then return the adapter. However, knowledge distillation~\cite{sanh2019distilbert,hinton2015distilling} is very expensive, limiting its application. (c)~The proposed ScaleOT introduces a layerwise importance-aware compression method Dynamic LayerReplace, providing privacy-utility-scalable emulators for downstream task tuning. }
\label{fig:compare}
\end{figure*}

Recent years have witnessed the bloom of large language models (LLMs)~\cite{bert,gpt2,llama,du2021glm}. Pre-trained on massive datasets, these models acquire robust general capabilities to tackle a comprehensive range of tasks by zero-shot (ZS) predictions or in-context learning~\cite{dong2022survey, wies2024learnability} and able to be fine-tuned for complex downstream tasks~\cite{wortsman2022robust,zhou2022conditional, wei2021finetuned,ouyang2022training,tans,liu2024visual,dai2024instructblip}. However, conventional centralized fine-tuning processes require the model and the data to be co-located - either the data owner uploads the data, which may threaten the data owner's data privacy, or the model owner shares the model weights - posing a risk of leaking the expensively trained models~\cite{chua2023fedpeat, OT}. Moreover, the possible exposure of parameters for the latter may increase the susceptibility of their fine-tuned models to be attacked ~\cite{li2020bert,zou2023universal,bdatk,liu2024lora}. All the above issues may hinder the long-run sustenance of LLMs. 

To protect both model ownership and data privacy effectively, the concept of offsite-tuning~\cite{OT, chua2023fedpeat,crash} has been proposed. As shown in Fig.~\ref{fig:compare}(b), instead of training with the full model, it enables the data owner to conduct fine-tuning using a lossy compressed emulator provided by the model owner, whose paradigm would result in a performance-degraded emulator for data owners. The trained adapter is then returned to the model owner and plugged into the full model to create an adapted model with high performance. Particularly, as the key factor for model privacy, the performance discrepancy between the data owner's and owner's sites encourages downstream users to use the adapted full model. Thus, the primary challenge of offsite-tuning lies in efficiently compressing LLM to protect model privacy by maintaining the performance discrepancy but improving the adapted full model. 

Following the offsite tuning strategy, vanilla method OT~\cite{OT} employs an \textit{Uniform LayerDrop}~\cite{layerdrop} that uniformly drops a subset of layers from the full model, as shown in Fig.~\ref{fig:banner}(a). However, despite many parameters in large models being redundant~\cite{michel2019sixteen}, the importance of each layer varies significantly~\cite{yao2024layer}, and such uniform drop may lead to performance degradation of the adapted full model. Moreover, the direct LayerDrop causes a misalignment between the input and output hidden spaces of the dropped layers, which also leads to a decline in the performance of the owner site. Although knowledge distillation~\cite{sanh2019distilbert,hinton2015distilling} can alleviate this issue, the massive cost of training a required emulator at least half the size of the LLM means that the massive cost of training raises a significant drawback for providing emulators with varying compression ratios. 

To address these issues, this paper introduces a novel offsite-tuning framework, named ScaleOT, that provides lossy compressed emulators with various scales for model privacy and facilitates lossless tuning compared with full fine-tuning. As shown in Fig.~\ref{fig:compare}(c), our framework consists of two stages: importance estimation and emulator generation. For the first stage, we propose an importance-aware layer-replace-based algorithm \textit{Dynamic LayerReplace} (see Fig.~\ref{fig:banner}(c)) that involves a reinforcement learning method to determine the importance of each layer within LLM. Simultaneously, for the less important layers, a set of trainable harmonizers, which are lightweight networks for better alignment between the remaining layers, are dynamically selected and trained as substitutions. In the second stage, according to the learned importance scores, the original model layers and their corresponding harmonizers can be combined variously to generate emulators while maintaining satisfactory performance at the model owner's site, as shown in Fig.~\ref{fig:banner}(d). Empirically, we unveiled that using a rank-decomposition to further compress the remaining model layers can better certify privacy protection with mere modal performance degradation, from which we propose the Selective Rank Compression (SRC) method. 

Extensive experiments on multiple models and datasets show that 
our approach outperforms the previous approaches while adjusting the compressed emulator model size and the rank reduction ratio in SRC, validating its effectiveness and feasibility. 

In a nutshell, our contributions are threefold:
\begin{itemize}[leftmargin=*]
\item \textbf{A flexible approach to produces various-sized compressed models for offsite-tuning:} We introduce an importance-aware lossy compression algorithm, Dynamic LayerReplace, oriented for offsite-tuning with LLMs that can scale up emulators by reinforcement learning and the harmonizers. All these components enable the flexibility of producing compressed models in various sizes. 
\item \textbf{Further compression for better privacy with mere fine-tuning degradation:} We propose the Selective Rank Compression strategy that can further enhance the model privacy with minimal performance loss.
\item  Comprehensive experiments show that our proposed ScaleOT outperforms the state-of-the-art methods.
\end{itemize}

\section{Related Work}
\noindent\textbf{Large Language Models.}
The rapid development of hardware and the abundance of data available in the information age has made it possible to train large foundational models for various tasks, such as GPT-3~\cite{gpt3}, CLIP~\cite{clip},  SAM~\cite{sam}, etc. These models are designed to have powerful general capabilities, capable of broadly solving problems through zero-shot learning or in-context learning~\cite{dong2022survey, wies2024learnability}. However, when facing downstream tasks, transfer learning with a small set of data is the mainstream choice because it not only offers better performance but also avoids the extensive resources required for training models from scratch. Although the open-sourcing of many models has significantly advanced academic progress, the parameter exposure increases the risk of adversarial attacks~\cite{li2020bert,zou2023universal}. On the other hand, closed-source models~\cite{gpt3} present contradictions between model ownership and data privacy due to the co-location of data and model in conventional fine-tuning. In summary, addressing these pain points, offsite tuning can provide a privacy-preserving, secure, and sustainable solution.

\noindent\textbf{Offsite-tuning.}
Previous works mainly focus on protecting user data privacy, such as federated learning~\cite{nguyen2021federated}. However, the widespread usage of LLMs raises concerns about the ownership of the model, considering the significant cost and time required to train large models. Therefore, how to protect model privacy has emerged as a research hotspot. OT~\cite{OT} was proposed as an effective bidirectional privacy-preserving solution for both model and data owners. Owing to its efficiency, it has been integrated into federated learning frameworks to facilitate privacy-preserving federated fine-tuning~\cite{chua2023fedpeat,fan2023fate,kuang2023federatedscope}. CRaSh~\cite{crash} improved OT by generating emulators by clustering, dropping and sharing layers. Additionally, some studies~\cite{dpopt} have explored the utilization of smaller language models to generate discrete prompts for offsite prompt tuning of large language models, though ensuring performance remains challenging. 

\noindent\textbf{Model Compression.}
The discovery of sparse sub-networks in convolutional, fully connected and Transformer models~\cite{lv2023lightformer,murty2211characterizing} has been validated by recent studies~\cite{hoefler2021sparsity,frankle2018lottery}. Previous works have found that models can be greatly compressed (often removing over 90\% of parameters) with very little drop in accuracy. However, conventional compression methods like pruning~\cite{deepcompression} and quantization~\cite{Jacob2018quan,radford2023robust}, while effective in compressing models, have been found to be difficult to use in offsite tuning~\cite{OT}, mainly due to consistent performance loss in both model and data sites. LayerDrop~\cite{layerdrop}, another explored method, although useful, can degrade the final performance of the adapted model, unless applying knowledge distillation. In this study, we propose a layer-replace-based compression method oriented for offsite tuning.

\section{Preliminary}
\label{sec:pre}
In our study, we consider the privacy concerns preventing the sharing and co-location of data and LLMs between their respective owners. Our objective is to tune the model using the data owner's data without accessing the model owner's model weights. Starting with a pretrained LLM $\mathcal{M}$, parameterized by weights $\Theta$, and a downstream dataset $\mathcal{D}$, we fine-tune this model on the downstream data to achieve $\mathcal{M}_\Theta \rightarrow \mathcal{M}_{\Theta+\Delta}, \Delta=\arg\min_{\delta}\mathcal{L}(\Theta+\delta,\mathcal{D})$. Our objective is to facilitate private transfer learning by finding an alternative, smaller, and weaker model $\mathcal{M}^*_{\Theta^*}$ (referred to as an Emulator) than $\mathcal{M}_\Theta$. This approach ensures that sharing $\mathcal{M}^*$ with downstream users does not threaten the ownership of the LLMs.
Then, data owners fine-tune the substitute model with their datasets, resulting in $\mathcal{M}^*_{\Theta^*+\Delta^*}$. 
We hope that by reintegrating the trained weights $\Delta^*$ into the original model (represented as $\mathcal{M}_{\Theta+\Delta^*}$), we can nearly mirror the performance observed when directly optimizing $\mathcal{M}$ on the dataset (denoted as $\mathcal{M}_{\Theta+\Delta}$), thereby eliminating the need to access to $\mathcal{M}$ itself.

For convenience, we make definations as follows: zero-shot (ZS) performance with $\mathcal{M}_{\Theta}$, 
fine-tuning (FT) performance with $\mathcal{M}_{\Theta+\Delta}$, emulator ZS and FT performance with $\mathcal{M}^*_{\Theta^*}$ and $\mathcal{M}^*_{\Theta^*+\Delta^*}$, and 
plug-in performance with $\mathcal{M}_{\Theta+\Delta^*}$ . An effective offsite-tuning should satisfy:
1)~ZS $<$ Plug-in so that necessitates the fine-tune process.
2)~Emulator FT $<$  Plug-in  to discourage the downstream users to use the fine-tuned emulators.
3)~Plug-in $\approx$ FT to encourage the downstream users to use the plugged model.

\section{Main Methodology}
\noindent\textbf{Problem setting.} In this paper, we concentrate on the design of offsite-tuning for the Transformer architecture~\cite{transformers}, which is extensively used in LLMs~\cite{gpt2,gpt3,llama}. We consider each transformer layer as the basic unit, and a LLM can be characterized as $\mathcal{M} = \{m_1,m_2,\dots,m_n\}$, where $n$ is the total number of the layers. Our approach involves dividing $\mathcal{M}$ into two components: a compact, trainable adapter $\mathcal{A}=\{m_i\}_{i \in \phi_{\mathcal{A}}}$ and the  remaining part of the model $\mathcal{E}=\{m_i\}_{i \in \phi_{\mathcal{E}}}$, making $\mathcal{M} = [\mathcal{A}, \mathcal{E}]$.
The sets of layer indices are defined such that $\phi_{\mathcal{A}} \cap \phi_{\mathcal{E}} = \emptyset$ and $\phi_{\mathcal{A}} \cup \phi_{\mathcal{E}} = \{1, 2, \ldots, n\}$. To safeguard the privacy of the model, a lossy compression is conducted on the frozen component $\mathcal{E}$, producing an emulator $\mathcal{E}^*$, thereby facilitating model tuning through updates to $\mathcal{A}$. After training at the data owner site, the updated adapter $\mathcal{A}'$ is returned to model owner side and replace the original $\mathcal{A}$  in the $\mathcal{M}$. Hence, the final updated LLM is denoted as $\mathcal{M}'=[\mathcal{A}',\mathcal{E}]$. Notably, the lossy compression inherently limits the model performance of $[\mathcal{A}',\mathcal{E}^*]$ for the downstream users, enabling the protection of model ownership. 

This paper tackles the two crucial keys of the problem: obtaining suitable division of $\mathcal{A}$ and $\mathcal{E}$ and achieving better compression from $\mathcal{E}$ to $\mathcal{E}^*$ for the effective fine-tuning and maintaining the privacy protection. For the former, we introduce the importance scores on the model layers which could be used to guide the selection of $\mathcal{A}$ and $\mathcal{E}$. In particular, the importance scores are estimated through reinforcement learning during the dynamically replacing  the original layers with lightweight networks. Those  lightweight networks, termed as harmonizers, can be further adopted as the substitutions for the layers in the $\mathcal{E}$, which promotes the performance of full adapted model. Moreover, for the remained layers in  $\mathcal{E}$ that replaced by the harmonizers,  we additionally propose the selective rank compression method, which certify better privacy while maintaining the full adapted model's performance. The following parts of the section introduce details of each proposed components. 

\subsection{Importance-aware Dynamic LayerReplace}
We propose a novel layer-replace-based compression algorithm called Dynamic LayerReplace. Our goal is to estimate the importance of each layer within LLM, and replace the less important layers with lightweight networks, named harmonizers, to preserve semantic consistency between layers.
To accomplish this, we utilize a dual-process approach comprising reinforcement learning (RL) to assess the importance of each LLM layer and deep learning (DL) for training the harmonizers via gradient descent. These processes alternate iteratively during the training phase to maintain stability. 

Formally, we begin with a LLM denoted as $\mathcal{M}$. We then initialize the importance scores $\mathcal{S} = \{s_1, s_2, \ldots, s_n\}$, and the harmonizers $\mathcal{H} = \{h_1, h_2, \ldots, h_n\}$. Two subsets of the dataset intended for pre-training are taken as the training sets $\mathcal{D}^T$ and validation sets $\mathcal{D}^V$, which are unrelated to the downstream tasks. During the training process, we update $\mathcal{S}$ utilizing RL and train $\mathcal{H}$ via DL, while keeping $\mathcal{M}$ unchanged. In the following, we will introduce LayerReplace sampling, which is the basic action of RL, and describe how to attain the importance score.

\noindent\textbf{LayerReplace Sampling.} We start by defining the state space for the RL process as the configuration of layers within the network, which includes both the presence of original layers and harmonizers. The decision to replace a specific layer with the corresponding harmonizer acts as the action, influenced by an action policy $\pi_i$ based on the importance score of each layer:
\begin{equation}
\label{eq:2}
    \pi_i \triangleq 
    U(0,\texttt{Sigmoid}(s_i)),
\end{equation}
where $U(a,b)$ denotes the uniform distribution between $a$ and $b$. For each time, we randomly sample a probability $p_i \sim \pi_i$, forming the probability set $\mathcal{P}=\{p_1,p_2,\dots,p_n\}$ for all layers. 
Empirically, half of the layers in LLMs are sampled following $\mathcal{P}$  and then replaced  with harmonizers. However, since the LLMs are typically deep, and the action policy in the early stages of training is inaccurate, selecting half of all layers directly risks selecting a significant number of adjacent layers, potentially leading to training collapse. To counter this and ensure training stability, we regroup the network's layers into $N_g$ groups of index of adjacent layers and replace half of the layers in each group. The set of the group is denoted as $\{ g_j\}_{j=1}^{N_g}$. The index set for the kept layers is constructed as follows:
\begin{equation}
\label{eq:3}
\phi = \{i \text{ where } p_i > p^{g_j}, i \in g_j \},
\end{equation}
where $p^{g_j}$ is the median probability in the $j$-th group, $N_g$ is empirically set to 4 by default. According to $\phi$, the sampled LayerReplace candidate network can be formed as:
\begin{equation}
F = f_1\circ f_2 \circ \dots \circ f_n,  \quad f_i = \left\{\begin{array}{ll}
m_i & \text{if } i \in \phi \\
h_i & \text{otherwise}
\end{array} \right. ,
\end{equation}
where $f_i\circ f_{i+1}$ represents compositional function. 

\noindent\textbf{Importance and Harmonizer Updating.}
We propose jointly updating importance scores with harmonizer to improve efficiency, involving training for DL and RL.
For the training of DL, we perform LayerReplace Sampling for once and update the parameters of harmonizers within the sampled candidate network through task loss using training dataset $\mathcal{D}^T$, i.e., $\theta_\mathcal{H}\leftarrow \nabla_{\theta}L(\theta)$. Here, we uses negative log-likelihood loss, a widely accepted standard for next-token prediction. Subsequently, the gradients necessary for updating the harmonizer are derived through backpropagation.

For the training of RL, we samples $N_c$ LayerReplace candidate networks based on the sampled probability sets $\{\mathcal{P}_j\}_{j=1}^{N_c}$. Then, we generate the index sets $\{\phi_j\}_{j=1}^{N_c}$, and their corresponding losses $\{\mathcal{L}_j\}_{j=1}^{N_c}$ with the held-out validation dataset $\mathcal{D}^V$. Last, the reward for $j$-th action policy can be written as:
\begin{equation}
    r_j = e^{-\mathcal{L}_j} -  \frac{1}{N_c}\sum\nolimits_{t=1}^{N_c}e^{-\mathcal{L}_t}.
\end{equation}
If the reward is greater than 0, it indicates that this policy is more optimal due to experiencing a smaller loss with the sampled candidate network, compared to other policies. 
Accordingly, the layers contained within the policy are more important. Therefore, the importance score is updated by:
\begin{equation}
\label{eq:6}
s_i = \left\{
\begin{array}{ll}
s_i + r_j*\sigma(s_i)*(1-\sigma(s_i)) & \text{if } i \in \phi_j \\
s_i & \text{otherwise}
\end{array}
\right. ,
\end{equation}
where $\sigma$ represent sigmoid function.

After training with Dynamic LayerReplace, we can obtain the importance scores for each layer of the LLM and the harmonizers to replace these layers, which will be used in the subsequent process of emulator creation. 

\subsection{Selective Rank Compression}
In many studies, LLMs have been found to hold an inherent over-parameterization nature~\cite{hoefler2021sparsity,hu2021lora}. Consequently, the model weights can be low-rank approximated without substantially impacting the model's performance. Intuitively,  we propose to compress the emulator's weights through low-rank approximation to enhance model privacy.  As the higher-order components of weights are reduced, the emulator's expressive capability is further diminished, enabling a larger performance gap. Meanwhile, the remaining lower-order components of weights still provide approximate gradient directions for updating adapters during tuning. In the following, we will briefly introduce the rank-$r$ approximation, and describe how we reduce the rank of specific layers for offsite tuning tasks.

\noindent\textbf{SVD-based Rank-$r$ Approximation. }
Given a matrix $W\in \mathbb{R}^{m\times n}$, and $r \in \mathbb{N}$, a rank-$r$ approximation problem seek to find a matrix $\hat{W}$ that minimizes $\|W-\hat{W} \|_2$ while ensuring that the rank of $\hat{W}$ is at most $r$. Singular Value Decomposition (SVD) is proven optimal to this problem according to Eckart-Young-Mirsky theorem~\cite{eckart1936approximation}. To achieve the rank-$r$ approximation using SVD, the matrix $W$ is decomposed as follows:
\begin{equation}
    W = U\Sigma V^T,
\end{equation}
where $U$ and $V$ are orthogonal matrices containing the left and right singular vectors of $W$, respectively, and $\Sigma$ is a diagonal matrix with the singular values of $W$ in descending order.

To form the rank-$r$ approximation $\hat{W}$, only the first $r$ singular values and corresponding singular vectors are used:
\begin{equation}
   \hat{W} = U_r \Sigma_r V_r^T .
\end{equation}
Here, $U_r$ and $V_r$ consist of the first $r$ columns of $U$ and $V$, and $\Sigma_r$ includes only the top $r$ singular values.

\begin{figure}[t]
\centering
\includegraphics[width=1\linewidth]{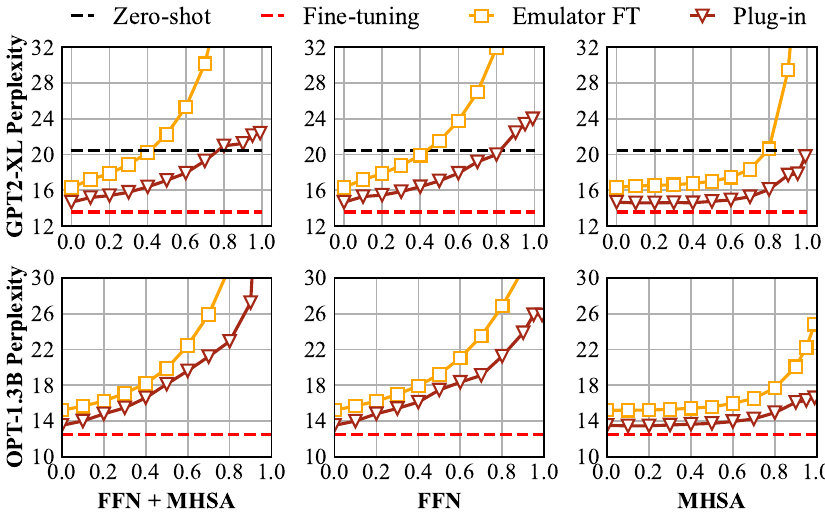}
\caption{Rank Compression Study with varies $\beta$ on multi-head self-attention (MHSA) layer and a Feedforward Network (FFN) layer in Transformer Block on WikiText dataset. }
\label{fig:src}
\end{figure}

\noindent\textbf{Rank Compression on Specific Modules. }
A typical Transformer~\cite{transformers} layer is composed of a multi-head self-attention (MHSA) layer followed by a Feedforward Network (FFN) layer. The MHSA layers facilitate interactions among tokens, while the FFN layer further processes information transformation within tokens. For expressive capability, the FFN's hidden dimension is empirically set to be very high, significantly exceeding its input and output dimensions (around $2.5\times$ to $4\times$). Considering that FFNs could be inherently high-rank, we leverage this characteristic and propose selectively applying rank compression to the weights in the MHSA layers to enhance the model privacy of the emulator. Empirical experiments, as shown in Fig.~\ref{fig:src}, indicate that applying rank compression to the weights in all layers (MHSA + FFN) and to the FFN alone both exponentially degrade the performance of both the model and data sites. In contrast, rank compression of the weights in MHSA leads to a rapid decline in the performance of emulator FT but a slow decrease in the performance of plug-in, especially when the rank compression ratio is larger than 0.6. Therefore, we choose to rank compress MHSA layers in the emulator to  enhance model privacy, thus form Selective Rank Compression strategy.

\begin{table*}[t]
\centering
\small
\begin{tabular}{cccccccccccc}\toprule[1.5pt]
\textbf{Method }& \textbf{Setting} & \textbf{OBQA} & \textbf{PIQA} & \textbf{ARC-E} & \textbf{ARC-C} & \textbf{HellaSwag} & \textbf{SciQ} & \textbf{WebQs} & \textbf{RACE} & \textbf{Avg.} & \textbf{$\Delta\uparrow$} \\\hline
\multicolumn{12}{c}{\textbf{GPT2-XL}} \\\hline
\multirow{2}{*}{Full Model} & Zero-shot (ZS) & 23.0 & 70.9 & 58.2 & 25.1 & 40.0 & 83.2 & 1.5 & 33.0 & 41.9 & \multirow{2}{*}{-} \\
 &Fine-tuning (FT) & 30.0 & 73.2 & 62.9 & 30.0 & 40.7 & 92.5 & 26.4 & 43.2 & 49.9 &  \\\hdashline
\multirow{3}{*}{OT} &  Emulator ZS $\downarrow$&18.8 &67.7 &53.2 &20.8 &33.5 &77.0 &0.2 &30.0 & \underline{37.7}  & \multirow{3}{*}{ \underline{2.9}} \\
 &  Emulator FT $\downarrow$& 24.0 & 70.3 & 58.2 & 23.9 & 35.8 & 92.7 & 18.9 & 39.4 &  \underline{45.4} & \\
 & Plug-in $\uparrow$& 28.2 & 73.6 & 61.4 & 28.5 & 41.6 & 93.2 & 19.9 & 39.9 & 48.3 &  \\\hdashline
\multirow{3}{*}{\begin{tabular}[c]{@{}l@{}}ScaleOT\\ w/o SRC\end{tabular}} & Emulator ZS $\downarrow$ & 19.4 & 69.5 &52.9 &23.9 &36.6 &81.3 &1.2 &33.3 &39.8  & \multirow{3}{*}{1.6} \\
 & Emulator FT $\downarrow$ & 29.6 & 72.7 & 58.6 & 26.5 & 39.6 & 93.2 & 24.4 & 43.4 & 48.5 &  \\
 &  Plug-in $\uparrow$  & 31.2 & 73.6 & 63.5 & 29.4 & 41.8 & 93.9 & 24.4 & 42.6 & \textbf{50.1} &  \\\hdashline
\multirow{3}{*}{ScaleOT} & Emulator ZS $\downarrow$ &15.4&61.3&37.9&19.5&28.6&53.8&0.0&24.2&\textbf{30.1} & \multirow{3}{*}{\textbf{4.3}} \\
& Emulator FT $\downarrow$ & 29.8 & 70.9 & 54.4 & 23.0 & 35.7 & 90.7 & 16.3 & 38.9 & \textbf{45.0} &\\
 &  Plug-in $\uparrow$  & 31.6 & 73.4 & 63.6 & 29.1 & 41.3 & 94.2 & 21.2 & 40.3 &  \underline{49.3} &  \\\hline\hline
\multicolumn{12}{c}{\textbf{OPT-1.3B}} \\\hline
\multirow{2}{*}{Full Model} &Zero-shot (ZS) & 23.4 & 71.6 & 56.9 & 23.5 & 41.5 & 84.4 & 4.6 & 34.2 & 42.5 & \multirow{2}{*}{-} \\
 & Fine-tuning (FT) & 31.4 & 75.2 & 61.3 & 27.7 & 42.7 & 92.5 & 31.2 & 37.0 & 49.9 &  \\\hdashline
\multirow{3}{*}{OT} & Emulator ZS $\downarrow$ & 19.4 &68.7 &53.9 &21.5 &35.1 &80.9 &1.3 &33.0 & \underline{39.2}  & \multirow{3}{*}{2.4} \\
 & Emulator FT $\downarrow$ & 24.8 & 71.6 & 58.1 & 26.1 & 37.0 & 92.2 & 24.3 & 38.6 &  \underline{46.6} & \\
 &  Plug-in $\uparrow$  & 29.0 & 74.5 & 59.4 & 27.8 & 43.3 & 92.9 & 26.2 & 38.9 & 49.0 &  \\\hdashline
\multirow{3}{*}{\begin{tabular}[c]{@{}l@{}}ScaleOT\\ w/o SRC\end{tabular}} &  Emulator ZS $\downarrow$& 20.2 &69.8 &54.1 &24.6 &38.2 &85.2 &1.4 &36.3 &41.2  & \multirow{3}{*}{ \underline{3.0}} \\
 &  Emulator FT $\downarrow$& 28.4 & 71.0 & 52.9 & 27.5 & 38.7 & 90.9 & 27.3 & 39.9 & 47.1 & \\
 &  Plug-in $\uparrow$  & 29.6 & 74.5 & 61.6 & 27.9 & 43.7 & 93.3 & 29.8 & 41.9 & \textbf{50.3} &  \\\hdashline
\multirow{3}{*}{ScaleOT} & Emulator ZS $\downarrow$& 17.2 &63.1 &41.8 &19.5 &32.1 &59.9 &0.1 &27.2 &\textbf{32.6}  & \multirow{3}{*}{\textbf{3.7}} \\
& Emulator FT $\downarrow$ & 27.2 & 70.9 & 52.5 & 26.5 & 37.8 & 90.0 & 25.3 & 39.1 & \textbf{46.2} & \\
 &  Plug-in $\uparrow$  & 28.2 & 75.2 & 61.9 & 28.3 & 42.9 & 94.0 & 28.2 & 40.8 &  \underline{49.9} & \\\bottomrule[1.5pt]
\end{tabular}
\caption{Comparative results of offsite-tuning with medium-size language model on eight question answering benchmarks. $\Delta$ denote the performance difference between Emulator fine-tuning and Plug-in. Best in \textbf{bold} and second best in \underline{underline}.}
\label{tab:medlan}
\end{table*}

\subsection{Privacy-utility-scalable Emulator Creation}
The creation of a privacy-utility-scalable emulator is defined by three quantities $(N_a, \alpha, \beta)$, consisting of the number of the adapted layers $N_a$, the proportion of layers replaced with harmonizers within LLM $\alpha$, and the rank reduction ratio of SRC $\beta$. Taken together, these values describe how the emulator $\mathcal{E}^*$ is created with LLMs $\mathcal{M}$, the important scores $\mathcal{S}$ and the harmonizers $\mathcal{H}$. 

Formally, given $\{ g_j\}_{j=1}^{N_g}$, we identify two sets of indices: 
\begin{align}
\phi_\mathcal{A}& =  \{i \text{ where } s_i \ge s^{g_j,k}, i \in g_j \}, \\
\phi_\mathcal{E}& =  \{i \text{ where } s_i < s^{g_j,k}, i \in g_j \},
\end{align}
where $k=\frac{N_a}{N_g}$, $ s^{g_j,k}$ is the $k$-th largest importance score in $j$-th group. Our goal is to tune the most important layers while keeping the less ones within the LLM unchanged, thus forming $\mathcal{A}=\{m_i\}_{i \in \phi_{\mathcal{A}}}$ and $\mathcal{E}=\{m_i\}_{i \in \phi_{\mathcal{E}}}$. Subsequently, the indices set of harmonizers is denoted as follows:
\begin{equation}
\phi_\mathcal{H} =  \{i \text{ where } s_i < s^{g_j,\kappa}, i \in g_j \},
\end{equation}
where $\kappa=\frac{n \times \alpha}{N_g}$, $s^{g_j,\kappa}$ is the $\kappa$-th largest importance score in $j$-th group, $\phi_{H} \in \phi_{E}$. The emulator $\mathcal{E}^*$ can be formed as:
\begin{equation}
    \mathcal{E}^*= \{ h_i\}_{i \in \phi_\mathcal{H} } \cup \left\{ \text{SRC}(m_i, \beta)\right\}_{i \in \phi_\mathcal{E} \land i \notin \phi_\mathcal{H}}.
\end{equation}

By modulating $\alpha$ and $\beta$, we effectively manage the model's privacy. Increasing values of $\alpha$ and $\beta$ enhances privacy via higher compression rates, while decreasing these values enhances the adapted model's performance. After tuning $\mathcal{A}$ with $\mathcal{E}^*$, the downstream users return $\mathcal{A}'$ to model owner, to form the well-tuned LLMs $\mathcal{M}'=[\mathcal{A}',\mathcal{E}]$.

\section{Experiments}
\subsection{Experimental Setup}
\noindent \textbf{Models and Datasets.} We evaluate our method on large language models, including GPT-2-XL~\cite{gpt2}, OPT-1.3B~\cite{opt}, OPT-6.7B~\cite{opt} and LLaMA~\cite{llama}. We validate our method across one generation task WikiText~\cite{wikitext}, and eight question answering benchmarks:  OBQA~\cite{openbookqa}, PIQA~\cite{piqa}, ARC~\cite{arc}, HellaSwag~\cite{hellaswag}, SciQ~\cite{sciq}, WebQuestions~\cite{webqs} and RACE~\cite{race}. 

\noindent \textbf{Evaluation Protocols.} For a fair comparison, we adopt the same evaluation metric used in previous studies~\cite{OT}. Specifically, we measure perplexity on the WikiText dataset and report accuracy on the other tasks. We use \texttt{lm-eval-harness}~\footnote{{https://github.com/EleutherAI/lm-evaluation-harness}} for language model evaluation.

\begin{figure}[t]
\centering
\includegraphics[width=.95\linewidth]{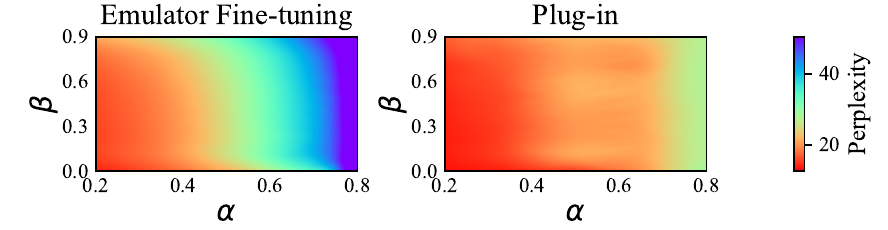}
\caption{Joint effect of $\alpha$ and $\beta$ in emulator generation.}
\label{fig:joint_sensitive}
\end{figure}

\noindent \textbf{Implementation Details.} In the training of the Dynamic LayerReplace, we utilize the Pile corpus~\cite{gao2020pile} datasets for language. For Dynamic LayerReplace, $N_c=3$ and $N_g=4$ are set empirically. For harmonizer, we utilize a simple low-rank FFN with ReLU activation and rank of 64 and 256 for medium and large size LLM respectively. For the construction of emulators, we set $\alpha = 0.25$ and $\beta = 0.8$ by default to balance the privacy-utility trade-off, unless otherwise specified. For fair comparison, $N_a$ is set to be consistent with OT~\cite{OT}, meaning that only about 10\% of the parameters are tuned, as opposed to full fine-tuning. For the offsite tuning phase, we employ the AdamW Optimizer, experimenting with a range of learning rates: [2e-5, 5e-5, 1e-4, 2e-4, 3e-4]. All experiments are conducted on a workstation with 8 V100 GPUs.

\begin{table*}[t]
\centering
\small
\begin{tabular}{rccccccccc}\toprule[1.5pt]
\multicolumn{1}{c}{\textbf{Setting}} & \textbf{OBQA} & \textbf{PIQA} & \textbf{ARC-E} & \textbf{ARC-C} & \textbf{HellaSwag} & \textbf{SciQ} & \textbf{WebQs} & \textbf{RACE} & \textbf{Avg.} \\\hline
\multicolumn{10}{c}{\textbf{OPT-6.7B}} \\\hline 
Zero-shot & 27.6 & 76.2 & 65.5 & 30.6 & 50.5 & 90.1 & 8.8 & 38.2 & 48.4 \\
Emulator ZS & 23.9 & 59.5 &58.1  & 26.4 & 30.9 &  86.6 & 2.0 & 33.4 & 40.0 \\
Emulator FT & 30.2 & 73.0 & 67.8 & 28.0 & 45.1 & 92.4 & 21.9 &44.0  &49.5  \\\hdashline
OT Plug-in & 33.8 & 77.7 & 66.8 & 33.9 & 52.1 & 91.9 & 23.9 & 44.1 & 53.0 \\
CRaSh Plug-in & 38.8 & 78.0 & 70.7 & 36.3 & 53.4 & 95.3 & 26.1 & 45.2 & 55.3 \\
ScaleOT w/o SRC Plug-in & 41.2 & 76.6 & 70.2 & 35.9 & 53.5 & 94.5 & 35.2 & 46.9 & \textbf{56.8} \\
ScaleOT Plug-in & 33.2 & 78.1 & 70.1 & 35.3 & 52.2 & 95.7 & 33.9 & 45.3 & \underline{55.5} \\\hline\hline
\multicolumn{10}{c}{\textbf{LLaMA-7B}} \\\hline
Full Zero-shot &28.2& 78.3 & 67.3 & 38.2 & 56.4 & 89.7 & 0.0 & 40.0 & 49.8 \\
Emulator ZS &23.6&72.4&57.8&29.9&46.1&81.5&0.0&35.6&43.4\\
Emulator FT & 31.0 & 74.7 & 62.3 & 28.8 & 44.2 & 92.3 & 21.8 & 42.9 & 49.8 \\\hdashline
OT Plug-in & 33.0 & 78.8 & 69.6 & 39.0 & 57.4 & 83.5 & 27.3 & 44.0 & 54.1 \\
CRaSh Plug-in & 34.6 & 80.0 & 71.3 & 41.8 & 58.4 & 95.1 & 29.8 & 45.6 & 57.1 \\
ScaleOT w/o SRC Plug-in & 40.8 & 79.6 & 75.3 & 47.7 &59.5  & 96.2 & 43.2 &  48.5& \textbf{61.2} \\
ScaleOT Plug-in & 37.4 & 79.7 & 73.2 & 42.3 & 58.1 & 95.7 & 33.7  & 45.6 & \underline{58.2} \\\bottomrule[1.5pt]
\end{tabular}
\caption{Comparative results of offsite-tuning with large-size language model on eight question answering benchmarks.  }
\label{tab:biglan}
\end{table*}

\subsection{Scalability in Emulator Generation.}
To showcase the scalability of privacy and utility using our proposed ScaleOT, we conducted experiments using OPT-1.3B with WikiText by tuning $\alpha$ and $\beta$. As demonstrated in Fig.~\ref{fig:joint_sensitive}, gradually increasing $\alpha$ causes a simultaneous decline in both emulator fine-tuning and plug-in performance. Meanwhile, increasing $\beta$ results in a larger performance discrepancy without harming plugin performance. We observed a broad parameter space to create a reasonable emulator, particularly when $\alpha$ is less than 0.5 and $\beta$ is greater than 0.6. This flexibility allows us to adjust our strategy to create either high-compressed, high-privacy emulators or low-compressed, high-performance emulators.

\subsection{Main Results}
We first evaluate our proposed ScaleOT on medium-sized models in Tab.~\ref{tab:medlan}, including GPT2-XL and OPT-1.3B with around one billion parameters. All methods meet the conditions of offsite tuning, i.e., the performance of the plug-in surpasses both the full model zero-shot and the performance of emulator fine-tuning. Moreover, ScaleOT without SRC achieves nearly lossless performance compared to full fine-tuning. This highlights the efficacy of Dynamic LayerReplace against the Uniform LayerDrop used in the baseline OT. Notably, due to the selection of important layers for updating, the performance of the plug-in can exceed that of directly fine-tuning on LLMs, thanks to the better convergence brought by sparse training~\cite{pan2024lisa,yao2024layer}. Finally, the incorporation of SRC significantly reduces the performance of emulator zero-shot and fine-tuning by average 9.2\% and 2.2\%, with minimal drop in the performance of the plugin. Overall, ScaleOT not only achieves better performance but also ensures good model privacy, reflecting the effectiveness of our method.

Subsequently, we validated the effectiveness of larger LLMs, including OPT-6.7B and LLaMA-7B, each with approximately 7 billion parameters. As shown in Tab~\ref{tab:biglan}, OT did not achieve satisfactory performance due to its inability to perform knowledge distillation on limited hardware. CRaSh improved performance through LayerSharing, but its lack of healing performance after compression resulted in suboptimal outcomes. In contrast, ScaleOT made the compression of large models feasible by requiring only about 1-2\% of the parameters to be trained during the compression phase. Notably, our method achieve strong plug-in performance on the WebQs task, where the zero-shot accuracy is zero, highlighting its potential for new downstream applications. Furthermore, ScaleOT achieved commendable results, demonstrating that its effectiveness is not restricted to specific model sizes. This makes ScaleOT a valuable strategy for enhancing offsite-tuning outcomes across models of varying scales.

\subsection{Analytical Study}
\noindent\textbf{Effect of Selective Rank Compression (SRC).}
To evaluate the effectiveness of SRC in improving model privacy, we conducted experiments on the WikiText dataset with GPT2-XL and OPT-1.3B. As depicted in Fig.~\ref{fig:src}, we linearly increase the compression ratio $\beta$ from 0 to 1, resulting in lower rank in the network. A consistent decreases in both emulator fine-tuning and plug-in performance were observed as $\beta$ increased, particularly in configurations that included the feed-forward network (FFN), where a linear relationship was evident. In contrast, for MHSA configurations, within a range of 0.6 to 1, emulator FT performance showed an exponential decline, whereas plug-in performance exhibited a linear reduction. This indicates that SRC has potential to enhance model privacy without degrading overall performance.

\begin{figure}[t]
\centering
\includegraphics[width=1\linewidth]{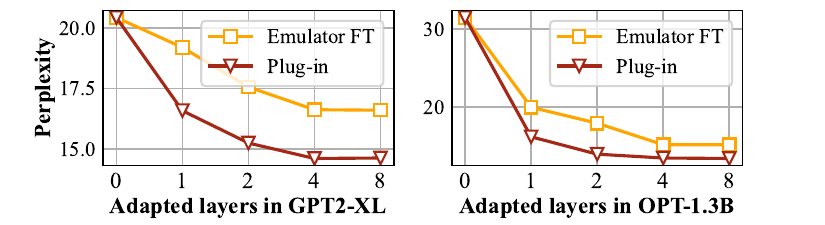}
\caption{Number of adapted layers in ScaleOT.}
\label{fig:apt}
\end{figure}

\noindent\textbf{Number of Adapted Layers.}
We analyzed the impact of the number of adapted layers $N_a$ on model performance. 
As illustrated in the Fig.~\ref{fig:apt}, With the increase in the number of adapted layers, the performance grows linearly and saturates at $N_a = 4$. Doubling $N_a$ to 8 does not yield any significant improvement. Consequently, $N_a = 4$ is selected for optimal balance between performance and computational efficiency.

\begin{figure}[t]
\centering
\includegraphics[width=1\linewidth]{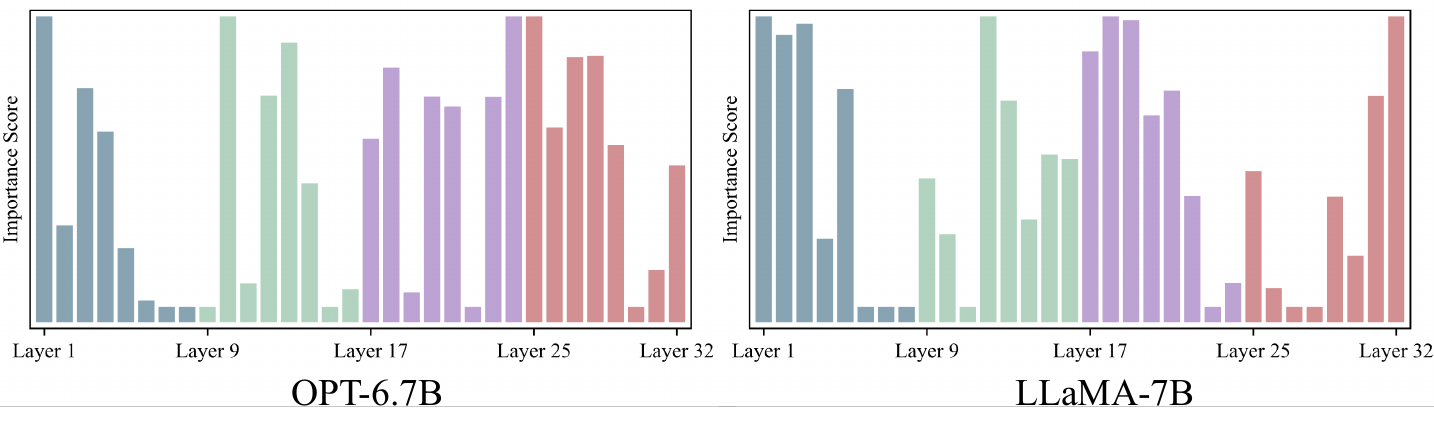}
\caption{Visualization of the estimated importance score.}
\label{fig:imp}
\end{figure}

\noindent\textbf{Importance Score.}
We visualize the estimated importance scores for OPT-6.7B and LLaMA-7B. As illustrated in Fig.~\ref{fig:imp}, it is evident that there is considerable variation in importance distribution across networks. However, a consistent pattern emerges: the first layer holds significant importance. This finding echoes the observations of OT~\cite{OT}, albeit lacking an explicit explanation. 

\begin{table}[t]
\small
\centering
\setlength\tabcolsep{3 pt}
\begin{tabular}{l|cccc}\toprule[1pt]
\multicolumn{1}{c}{Methods} & \begin{tabular}[c]{@{}c@{}}\#Total \\ Params\end{tabular} & \begin{tabular}[c]{@{}c@{}}\#Trainable \\ Params\end{tabular} & Emu. FT $\uparrow$ & Plug-in $\downarrow$ \\\midrule
Zero Shot & 1.64B & 0 & - & 20.44 \\
Full Finetune & 1.64B & 1.48B & - & 13.58 \\\hline
ScaleOT w/o SRC & \multirow{4}{*}{1.27B} & 127M & 16.62 & 14.60 \\
\quad+Adapter (d=64) &  & 1.65M &  17.29 & 14.92\\
\quad+LoRA (r=4) &  & 410K & 16.75 & \textbf{14.53}  \\\hline
ScaleOT & \multirow{4}{*}{1.05B} & 127M & 20.65 & 15.95 \\
\quad+Adapter (d=64) &  & 1.65M &  21.69 & 16.03\\
\quad+LoRA (r=4) &  & 410K & 21.18 & \textbf{15.85} \\\bottomrule[1pt]
\end{tabular}
\caption{ScaleOT is orthogonal to most existing parameter-efficient fine-tuning techniques. Scores are validated with GPT2-XL on WikiText-2 in terms of perplexity.}
\label{tab:lora}
\end{table}

\noindent\textbf{Orthogonality with Parameter-efficient Fine-tuning.}
ScaleOT is designed to integrate seamlessly with parameter-efficient fine-tuning (PEFT) methods, facilitating a combined approach that significantly decreases the trainable parameters and enhances efficiency. This can be accomplished by employing PEFT methods in adapter layers, including strategies such as Adapter-tuning~\cite{houlsby2019parameter} and LoRA~\cite{hu2021lora}. As shown in Tab.~\ref{tab:lora}, we observed that Adapter-tuning and LoRA substantially reduced the trainable parameters while maintaining plug-in performance. 

\section{Conclusion}
In this paper, we propose a novel privacy-utility-scalable offsite-tuning framework, namely ScaleOT. We introduce a layerwise importance-aware lossy compression algorithm, Dynamic LayerReplace, to scale up the emulator and adapter by reinforcement learning and the harmonizers. Then, we propose Selective Rank Compression (SRC) that can enhance model privacy with minimal impact on performance. Comprehensive experiments across various LLMs and tasks showcase the superiority of our proposed ScaleOT.

\section{Acknowledgments}
This work was supported by Ant Group Postdoctoral Programme.

\bibliography{aaai25}

\appendix

\section{Appendix}

\subsection{Connection to Memory-Efficient Fine-tuning.}
We have noticed the recently published work, LISA~\cite{lisapan}, closely related to ours, which considers layerwise importance to enhance memory-efficient fine-tuning. Specifically, they share a similar idea of training LORA adapters only on a subset of layers during fine-tuning, as we do. Their findings suggest that fine-tuning just a few layers can achieve better results than training LORA across all layers, providing empirical validation for our observation that $\mathcal{P}_{pl}>\mathcal{P}_{ft}$ for GPT2-XL and OPT-1.3B. However, despite they claim to employ an importance-aware methodology, they actually select a few layers at random (from 2 to 4) for each training iteration. In contrast, our approach explicitly estimates the importance of layers, serving as a superior alternative to their method. Consequently, we believe that our method could be generalized to Memory-Efficient Fine-tuning methods, potentially further enhancing the performance of LISA.

\subsection{Comparison of Zero-shot Performance of Emulator.}
\begin{table}[h]
\small
\setlength\tabcolsep{1 pt}
\centering
\caption{Zero-shot performance of emulator on GPT2-XL. Lower is better.}
\label{tab:zse}
\begin{tabular}{c|cccccccc|c}
\toprule[1pt]
Methods & OBQA & PIQA & ARC-E & ARC-C & HSwag & SciQ & WebQs & RACE & Avg$\downarrow$\\
\midrule
Raw & 23.0 & 70.9 & 58.2 & 25.1 & 40.0 & 83.2 & 1.5 & 33.0 & 41.9\\
OT & 18.8 &	67.7 &	53.2 & 20.8 & 33.5 &77.0 & 0.2 & 30.0 & 37.7\\
ScaleOT & 15.4 & 61.3 & 37.9 & 19.5 & 28.6 & 53.8 & 0.0 & 24.2 & 30.1\\
\bottomrule[1pt]
\end{tabular}
\end{table}

To illustrate the efficacy of our approach, we provide an extended evaluation of our lossy compressed emulator's zero-shot performance at the data owner's site, in contrast with the baseline OT algorithm. As depicted in Table~\ref{tab:zse}, the original model demonstrates a zero-shot performance with an average of 41.9. Despite the baseline OT algorithm attaining an average of 37.7, it maintains considerable zero-shot performance on particular datasets, including PIQA and RACE. ScaleOT, in comparison, records the emulator's lowest zero-shot performance at 30.1, thereby underscoring the effectiveness of our method in protecting model privacy.

\subsection{Limitation and Future Work}
Our method offers an appealing privacy-utility trade-off, where it can provide lossless offsite-tuning or the enhanced model privacy. However, our technique still requires training during the model compression process, which limits the cost-effectiveness of its application. In future work, we plan to explore lossy compression methods that do not require training to reduce the costs associated with offsite tuning, such as employing LayerSharing~\cite{lan2019albert} techniques to replace the LayerReplace strategy.

\end{document}